\documentclass[10pt,journal,compsoc]{IEEEtran}
\newif\ifpeerreview

\peerreviewfalse

\usepackage[nocompress]{cite}
\usepackage{url}
\usepackage{amsmath,amssymb,graphicx}

\usepackage{lipsum} 

\usepackage{paralist}
\usepackage[switch]{lineno}

\usepackage{tabularx}
\usepackage{booktabs}
\usepackage{multirow}
\usepackage{bbm}
\usepackage{amsmath}

\newcommand{\myparagraph}[1]{\vspace{0.1em}\noindent\textbf{#1}}

\newcommand{\paperID}{26}

\title{MirrorNeRF: One-shot Neural Portrait Radiance Field from Multi-mirror Catadioptric Imaging}

\author{Ziyu~Wang,
        ~Liao~Wang,
        ~Fuqiang~Zhao,
        ~Minye~Wu,\\
        ~Lan~Xu,
        and~Jingyi~Yu,~\IEEEmembership{Fellow,~IEEE}
\IEEEcompsocitemizethanks{\IEEEcompsocthanksitem Z.~Wang, L.~Wang, and F.~Zhao are with the School of Information Science and Technology, ShanghaiTech University, Shanghai, 201210, China. 
\protect\\E-mail: \{wangzy6,wangla,zhaofq\}@shanghaitech.edu.cn
\IEEEcompsocthanksitem M.~Wu is with the School of Information Science and Technology,
ShanghaiTech University, Shanghai 201210, China, and the Shanghai
Institute of Microsystem and Information Technology, Chinese Academy of
Sciences, Shanghai 200031, China, and also with the University of Chinese Academy of Sciences, Beijing, 100049, China.
\protect\\E-mail: wumy@shanghaitech.edu.cn.
\IEEEcompsocthanksitem L.~Xu, and J.~Yu are with the School  of  Information  Science and Technology, ShanghaiTech University, and also with the Shanghai Engineering Research Center of Intelligent Vision and
Imaging, Shanghai, 201210, China. 
\protect\\E-mail: \{xulan1,yujingyi\}@shanghaitech.edu.cn
}
} 

\begin{document}

\IEEEtitleabstractindextext{%
\begin{abstract}
Photo-realistic neural reconstruction and rendering of the human portrait are critical for numerous VR/AR applications. Still, existing solutions inherently rely on multi-view capture settings, and the one-shot solution to get rid of the tedious multi-view synchronization and calibration remains extremely challenging.
In this paper, we propose \textit{MirrorNeRF} -- a one-shot neural portrait free-viewpoint rendering approach using a catadioptric imaging system with multiple sphere mirrors and a single high-resolution digital camera, which is the first to combine neural radiance field with catadioptric imaging so as to enable one-shot photo-realistic human portrait reconstruction and rendering, in a low-cost and casual capture setting.
More specifically, we propose a light-weight catadioptric system design with a sphere mirror array to enable diverse ray sampling in the continuous 3D space as well as an effective online calibration for the camera and the mirror array. 
Our catadioptric imaging system can be easily deployed with a low budget and the casual capture ability for convenient daily usages.
We introduce a novel neural warping radiance field representation to learn a continuous displacement field that implicitly compensates for the misalignment due to our flexible system setting.
We further propose a density regularization scheme to leverage the inherent geometry information from the catadioptric data in a self-supervision manner, which not only improves the training efficiency but also provides more effective density supervision for higher rendering quality.
Extensive experiments demonstrate the effectiveness and robustness of our scheme to achieve one-shot photo-realistic and high-quality appearance free-viewpoint rendering for human portrait scenes.
\end{abstract}

}


\maketitle
\thispagestyle{empty} 

\begin{figure*}
	\centering 
	\centerline{\includegraphics[width=1\linewidth]{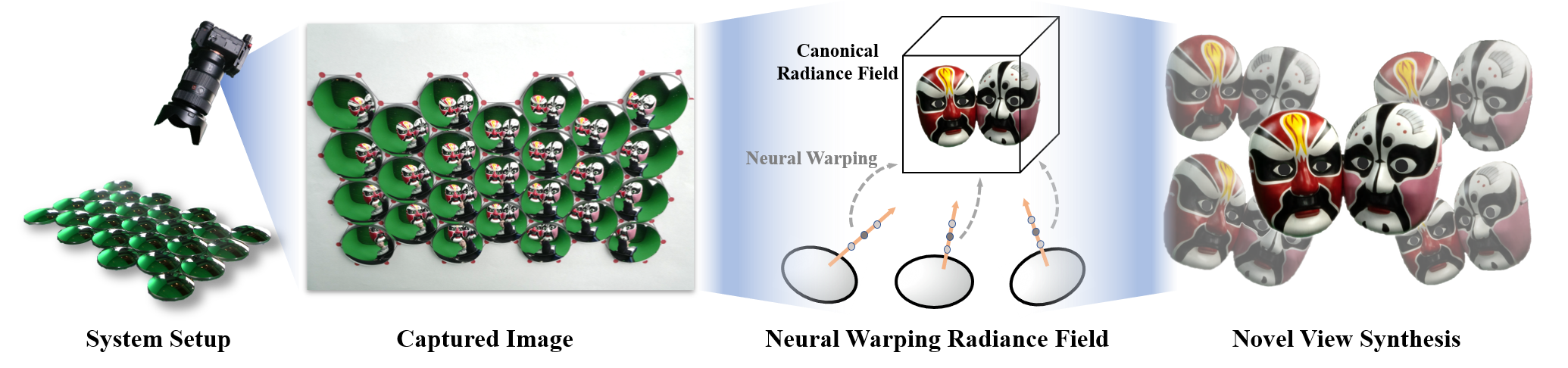}}
	\vspace{-10pt}
	\caption{Our MirrorNeRF approach combines a light-weight catadioptric imaging system with a novel neural scene representation {\it Neural Warping Radiance Field} (NeWRF), which achieves photo-realistic novel view synthesis using only a single camera shot.} 
	\label{fig:fig_1_teaser} 
\end{figure*}


\IEEEraisesectionheading{
  \section{Introduction}\label{sec:introduction}
}

\IEEEPARstart{W}{ith} the recent popularity of virtual and augmented reality (VR and AR) to present information in an innovative and immersive way, the high-quality 3D human-centric content generation evolves as a cutting-edge yet bottleneck technique.
Reconstructing a high-quality and photo-realistic human portrait scene conveniently for better VR/AR experience has recently attracted substantive attention of both the computer photography and computer graphics communities.

The recent data-driven neural rendering techniques~\cite{lombardi2019neural,Wu_CVPR2020,FreeViewSynthesis,sitzmann2019srns,mildenhall2020nerf} bring huge potential for realistic human portrait modeling and rendering in novel views using only RGB images as input.
Specifically, the recent approaches~\cite{mildenhall2020nerf,martin2020nerf} utilize neural radiance fields with volume rendering to achieve photo-realistic free-viewpoint results of complicated scenes.
However, these approaches require to dispatch camera rays through various view angles to model the continues plenoptic function space, which inherently relies on multi-view capture setting spatially~\cite{mildenhall2020nerf,zhang2020nerf++,liu2020neural} or temporally~\cite{ST_nerf,Deform_nerf,D_nerf}.  
Thus, reconstructing a one-shot neural radiance field for realistic portrait scenes through only a single camera remains extremely challenging.

In this paper, we propose a rescue to such problem by using a multi-mirror catadioptric imaging system.
The key idea is that the mirror is a natural dispatcher of rays to enable diverse sampling across the continues 3D space, which potentially allows 3D modeling of a scene through only one-shot capture.
Catadioptric imaging systems~\cite{baker1999theory,yu2004general,lanman2006spherical} consist of mirrors (cata) and lenses (dioptric) to capture a wide view of world with a single sensor and have been broadly used in various tasks such as panoramic imaging~\cite{krishnan2008cata}, 3D reconstruction~\cite{cata3D_2004CVPR,ding2009multiperspective,chen2019multi}, light-field modeling~\cite{levoy2004synthetic,taguchi2010axial} or hyperspectral sensing~\cite{xue2017catadioptric,baskurt2020catadioptric}.
Most relatively to our setting, the approaches~\cite{ding2009multiperspective,chen2019multi} enable one-shot 3D reconstruction by utilizing multi-stereo techniques with multiple spherical mirrors.
However, only coarse geometry can be recovered, leading to inferior texture results.
Another method~\cite{taguchi2010axial} utilizes similar spherical mirror array for wide-angle light-field rendering but still suffers from unrealistic blending artifacts due to the discontinuity of the recover geometry. 
More importantly, none of existing approaches explore to utilize catadioptric imaging to strengthen the neural radiance field generation, so as to provide photo-realistic reconstruction.
Moreover, to utilize the neural radiance field~\cite{mildenhall2020nerf}, these capture systems above require heavy manual labor to precisely calibrate the mirrors and the camera, leading to the high restriction of the daily usages with casually capture.

To tackle these challenges, we present \textit{MirrorNeRF} -- a one-shot neural rendering approach using a catadioptric imaging system with multiple sphere mirrors and a single high-resolution digital camera. 
As illustrated in Fig.~\ref{fig:fig_1_teaser}, our approach is the first to combine neural radiance field with catadioptric imaging, which enables one-shot photo-realistic human portrait reconstruction and rendering, whilst maintaining low-cost and casual capture setting.

More specifically, from the system side we first introduce a light-weight catadioptric imaging system using a sphere mirror array, where all the mirrors are arranged tightly in hexagons and form a honeycomb pattern to provide diverse sampling in the continues 3D space.
To enable easy deployment with a low budget, our hexagon design is drawn on a simple flat A3-size white paper with red dots on the vertices and all the mirrors are manually placed without tedious and absolutely precise arrangement.
An effective online calibration scheme based on the red dots for the mirror array and the camera is proposed to further enable casually one-shot capture.  
Compared with previous systems~\cite{lanman2006spherical,taguchi2010axial,chen2019multi}, our deployment is much more easy to setup with a low budget and casual capture ability, yet such convenience presents the challenges on the algorithm side to handle the misalignment caused by both the calibration and arrangement.	
From the algorithm perspective, we extend the neural radiance field into the catadioptric imaging.
Inspired by the work~\cite{mildenhall2020nerf} we first adopt the radiance representation and volumetric integration to obtain the density and color values along the rays dispatched by our catadioptric system.
To further handle the misalignment challenge due to our flexible system setting, we introduce a novel neural warping radiance field representation, which utilizes per-mirror latent code to learn a continuous displacement field, so as to implicitly compensate the complicated misalignment.
Moreover, a density regularization scheme is proposed to further leverage the inherent geometry information from the catadioptric data in a self-supervision manner, which not only improves the training efficiency but also provides more effective density supervision for higher rendering quality.
To summarize, our main contributions include:
\begin{itemize}
    \item We present a one-shot photo-realistic neural portrait rendering approach in novel views, which is the first to combine neural radiance field with multi-mirror catadioptric imaging.
    \item we propose a light-weight catadioptric system design to enable diverse ray sampling, which is low-cost, easily deployed and supports casual capture.
    \item We propose a neural warping radiance field representation as well as a density regularization scheme, which can handle the misalignment caused by our flexible system to enable high-quality rendering.
\end{itemize}

\section{Related Work}

\subsection{3D Scene Modeling}
Recent work has made a significant process on 3D object modeling and realism free-viewpoint rendering with level sets of deep networks that implicitly map spacial locations $xyz$ to geometric representations (i.g., distance field~\cite{park2019deepsdf}, occupancy Field~\cite{mescheder2019occupancy,chen2019learning} etc..).  In contrast to the aforementioned explicit representations which require discretization (e.g., in terms of the number of voxels, points, or vertices), implicitly models shapes with a continuous function and naturally is able to handle complicated shape topologies. The implicit geometric modeling can be easily learned from 3D point samples~\cite{peng2020convolutional,saito2019pifu,saito2020pifuhd}, and the trained models can be used to reconstruct shapes from a single image or 3D part. However, these models are limited by their requirement of access to ground truth 3D geometry, typically obtained from synthetic 3D shape datasets such as ShapeNet~\cite{chang2015shapenet}. Subsequent works relax this requirement of ground truth 3D prior by formulating differentiable rendering functions that allow neural implicit shape representations to be optimized using only 2D images~\cite{sitzmann2019srns,niemeyer2020differentiable}. 


When the viewing angle is zoomed in and out, there will be different degrees of voids or artifacts. Voxel occupancy provides a regular representation, but due to its memory-cost characteristics, it often only provide rough low-resolution object expression. In fact, our goal is to express the object in a continuous space that conforms to the physical facts. Therefore, we chose a continuous model of neural radiance field~\cite{mildenhall2020nerf}, as the basis of our work.

\subsection{Free-Viewpoint Rendering}
Free-viewpoint synthesis methods generally model input/target images as a collection of rays and essentially aims to recover the plenoptic function~\cite{Debevec1998Image} from dense samples. Previous Image-Based Rendering (IBR) works~\cite{Levoy1996Light} used two plane parametrization, or 2PP, to represent rays and render new rays via ray blending.
They are able to achieve real-time interpolation but require a lot of memory as they need to cache all rays. Following works Buehler \textit{et al.}\cite{Buehler2001Unstructured} bring in proxy geometric to select suitable views and filter occluded rays by cross-projection to the image plane when ray fusion. However, those methods are still limited by the linear blending function usually result in strong ghosting and blurring artifact. \cite{wood2000surface,chen2002light} seek to model the radiance field as rays emitted from the object surface, i.e., the surface light fields. The surface-based ray parameterization can achieve a more continuous ray interpolation and resolve the issue that the ray sample extremely unbalance in the previous 2PP modeling.

Most recently, seminal researches seek to implicitly represent the radiance field and render novel views with neural networks. Chen \textit{et al.}\cite{chen2018deep} presented Deep Surface Light Fields, which use an MLP network to fix per-vertex radiance and learns to fill up the missing data across angle and vertices. This ~\textit{et al.}~\cite{thies2019deferred} present a novel learnable neural texture to model rendering as image translation. 
Another line of research extends the free-viewpoint to dynamic sequence or scene relighting.
Wu~\emph{et al.}~\cite{Wu_CVPR2020} model and render dynamic scenes through embedding special features with the sparse dynamic point cloud. 
Lombardi~\emph{et al.}~\cite{lombardi2019neural} use a novel volumetric representation to reconstruct dynamic geometry and appearance variations jointly while requires only image-level supervision. 
Chen~\emph{et al.}~\cite{chen2020neural} model image formation in terms of environment lighting, object intrinsic attributes, and the light transport function (LTF) that achieve free-viewpoint relighting. Notable exceptions are the most recent Nerf~\cite{mildenhall2020nerf} and Nerf in the wild (Nerf-W)~\cite{martin2020nerf}. The Nerf implicitly models the radiance field and the density of a volume by neural networks, then uses direct volume render function to reconstruct geometric and novel views. 
The following work Nerf-W relaxes the strict consistency assumptions through modeling per-image appearance variations such as exposure, lighting, weather, and post-processing with a learned low-dimensional latent space. 
Our system combines neural radiance field with catadioptric imaging for the first time, which enables photo-realistic human portrait reconstruction and rendering within a single shot.

\subsection{Catadioptric Imaging}
Catadioptric imaging system consists of lenses and curved surfaces to achieve a wide field of view captured by a single sensor. 
Baker~\emph{et al.}~\cite{baker1999theory} present in detail the modeling and the configuration of single view point catadioptric image systems. 
These systems require an accurate alignment of the sensor optical axis and the focus of the curved mirrors to maintain the single viewpoint property. 
As for the non-single viewpoint catadioptric imaging system, there is no strict restriction on sensor location, making it possible for multi-mirror catadioptric system construction to realize multi-perspective capturing.
Levoy~\emph{et al.}~\cite{levoy2004synthetic} use a planer mirror array to capture light field.
Taguchi~\emph{et al.}~\cite{taguchi2010axial} propose a geometric non-approximate model for spherical mirror array light field capture. 
Xue~\emph{et al.}~\cite{xue2017catadioptric} propose spectral coded spherical mirror arrays for acquiring 5D light fields.
A multi-mirror catadioptric system can also be used for geometric reconstruction.
Lanman~\emph{et al.}~\cite{lanman2006spherical} manually select several corresponding points on each spherical mirror to recover vertex positions for the reconstruction of mesh.
Ding~\emph{et al.}~\cite{ding2009multiperspective} use piecewise GLCs~\cite{yu2004general} approximation for the reflective rays to enable stereo matching and fast projection for volumetric reconstruction.
Chen~\emph{et al.}~\cite{chen2019multi} propose a method for the multi-mirror system which enables multiple central and non-central compacted stereo matching for 3D reconstruction.

Differently, our system combines the catadioptric imaging with neural radiance field for the first time to our knowledge, which enables one-shot photo-realistic human portrait reconstruction and rendering whilst maintaining a low-cost and casual capture setting.

\section{Overview}
The proposed MirrorNeRF marries implicit neural radiance field with multi-mirror catadioptric imaging, which enables one-shot photo-realistic human portrait reconstruction and rendering in novel views.
Fig.~\ref{fig:fig_1_teaser} illustrates the high-level components of our approach, which takes a single catadioptric image captured by a digital camera as input, and generates high-quality novel-view synthesis results in various challenging human portrait scenarios as output.

\myparagraph{Catadioptric Imaging System.}
We first introduce a light-weight catadioptric system design to enable diverse ray sampling in one-shot capture, which is low-cost, easily deployed, and supports casual capture.
Our system consists of a sphere mirror array and a high-resolution digital camera, where all the mirrors are arranged tightly in hexagons and form a honeycomb pattern. (Sec.~\ref{Sec:capture_system}). 
%

\myparagraph{Neural Portrait Rendering.}
We then propose a neural portrait rendering scheme based on the catadioptric image with dispatched rays.
We adopt the radiance representation and volumetric integration of the work~\cite{mildenhall2020nerf} to obtain the density and color values along with these rays.
We introduce a neural warping radiance field representation which learns a continuous displacement field to handle the misalignment caused by our flexible system.
Moreover, a density regularization scheme is proposed to further leverage the inherent geometry information from the catadioptric data, which improves the training efficiency rendering quality (Sec.~\ref{Sec:neural_rendering}).


\section{Catadioptric Imaging System} \label{Sec:capture_system}
\begin{figure}[t]
	\centerline{\includegraphics[scale=0.43]{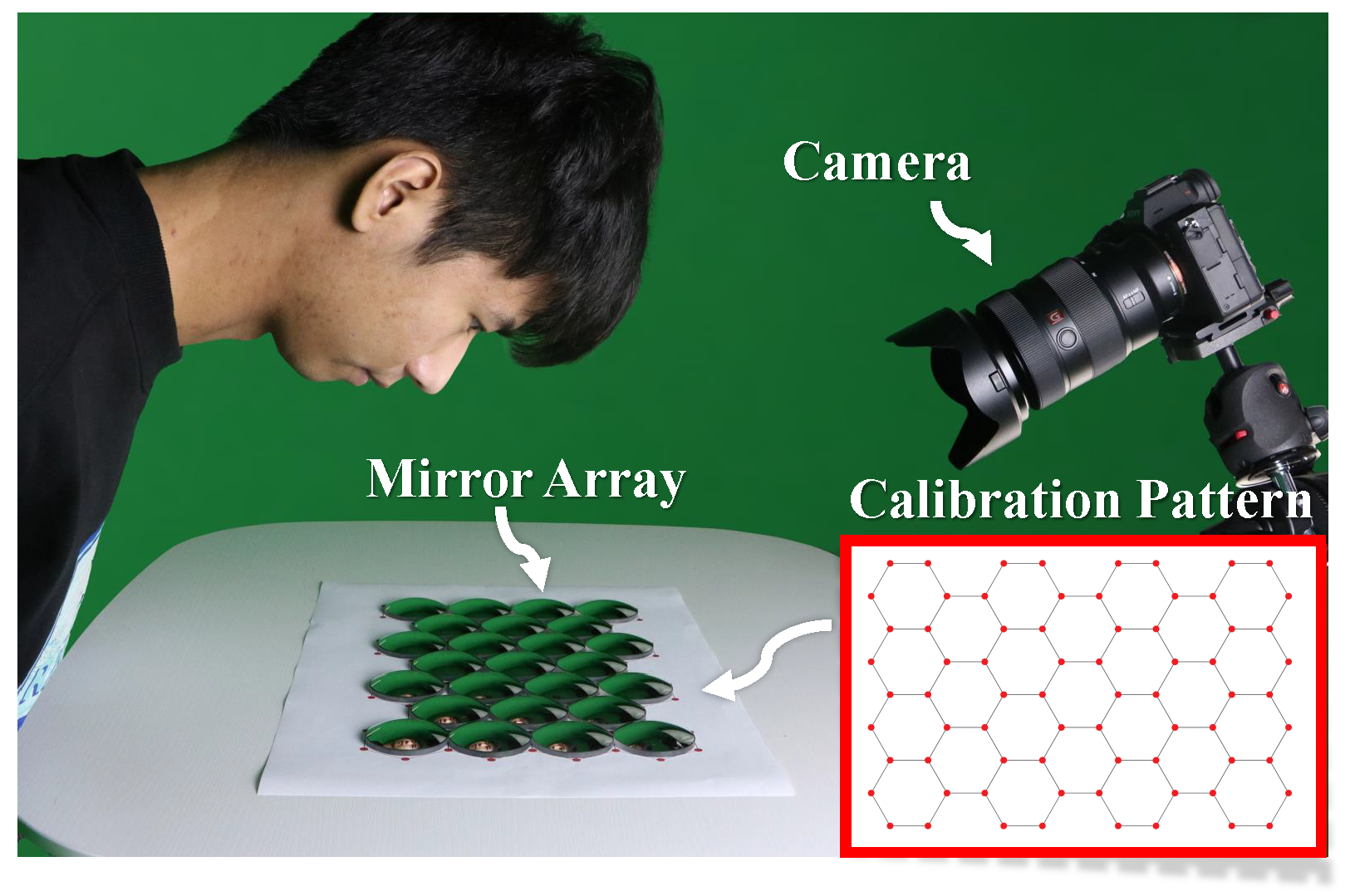}}
	\vspace{-10pt}
	\caption{Our capture system consists of a camera and a mirror array. The red box in the lower right corner shows the calibration pattern.}
	\label{fig:system}
\end{figure}

Our light-weight catadioptric system consists of a mirror arary and a digital camera, as shown in Fig.~\ref{fig:system}.
The camera faces toward the mirror array at a range of angles determined by the primitive shape of the mirror to capture the entire mirror array in one shot.
We will evaluate the angle range of our system setting in the experimental section. 
We use the high-resolution SONY ILCE7RM4 camera, which can take 61MP pictures for capturing more details. 
The system is decorated with a green screen as the background to facilitate easy foreground segmentation. 
During capturing, the target is placed in front of the mirror, and we shot only one image for reconstruction. 

\vspace{2mm}\noindent{\bf Mirror Array.} 
The mirror array in MirrorNeRF consists of 25 low-cost convex spherical reflecting mirrors. 
We scan the geometry of the mirror in advance to obtain its approximate parameters (e.g., diameter and thickness). 
Mirrors are placed tightly on a flat plane with an A3-size white paper. 
We draw hexagons on this paper and arrange them as a honeycomb pattern. 
Each hexagon has a similar size to the mirror and marks the mirror position. 
Thus, we can place a mirror over it manually without tedious and special precise operation. 
Such honeycomb pattern ensures that the captured rays have varieties of directions and span the continuous 3D space as much as possible for neural portrait reconstruction, as illustrated in Fig.~\ref{fig:system}. 

As described above, assembling this mirror array is convenient and requires no precise operations. 
We assume all the mirrors are put in the exact place and accordingly prepare a 3D geometry template of the whole mirror array as a 3D proxy for the following ray restoration process. 
Note that the actual geometry of the mirror array is inaccurate due to the manual placement since the misalignment between mirrors and hexagons violates our geometry assumption. %
We attack the misalignment challenge by introducing a warping field in the next section.


\begin{figure}[t]
	\centerline{\includegraphics[width=1\linewidth]{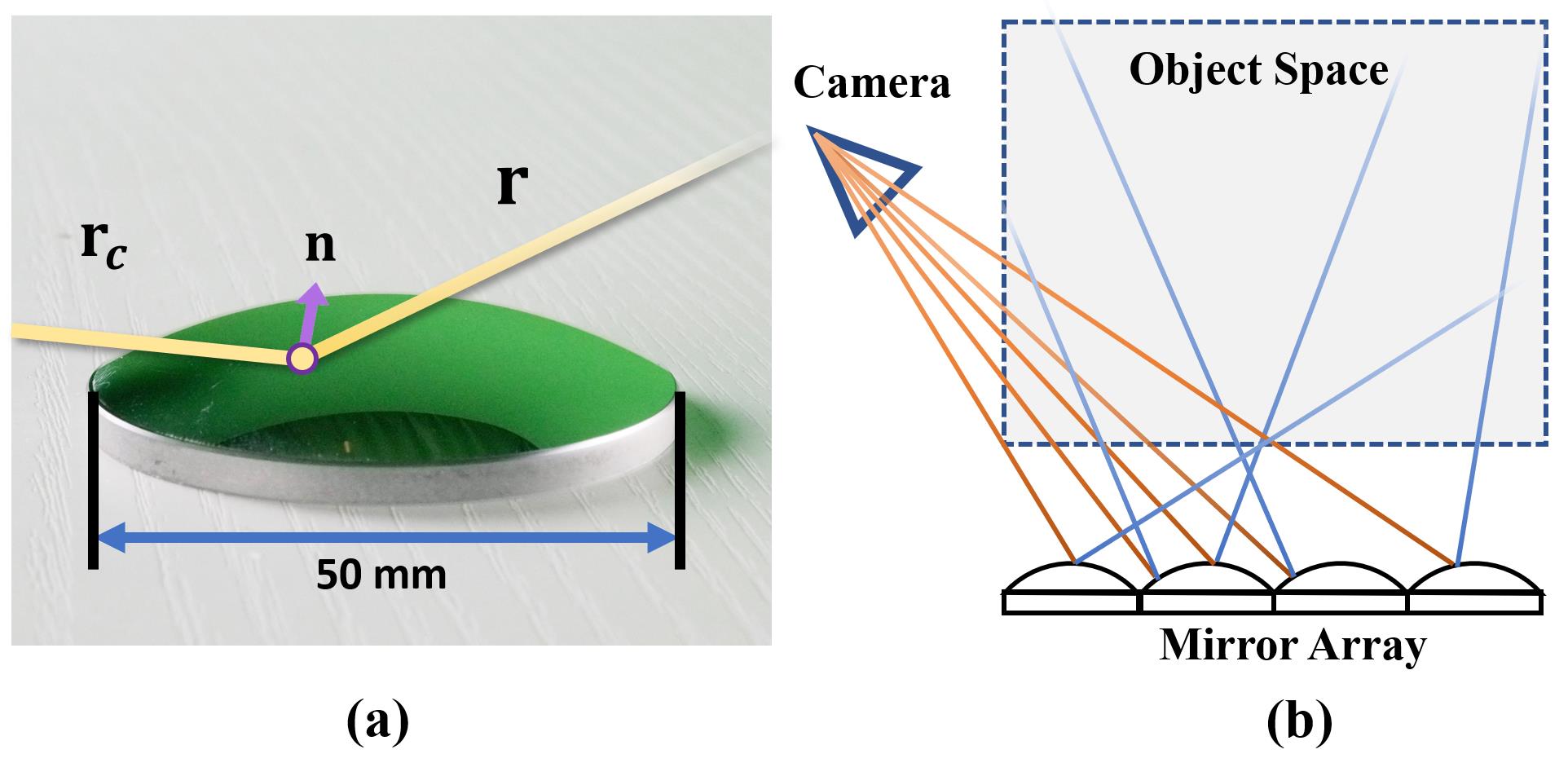}}
	\vspace{-16pt}
	\caption{Illustrations of light paths in our catadioptric system. (a) shows the reflection on a mirror surface; The mirror has a diameter of 50mm; (b) demonstrates that our system can capture rays coming from varying directions in a single image through a mirror array.  }
	\label{fig:mirrot_light_path}
\end{figure}

\begin{figure*}[t]
	\centering
    		\centerline{\includegraphics[width=1\linewidth]{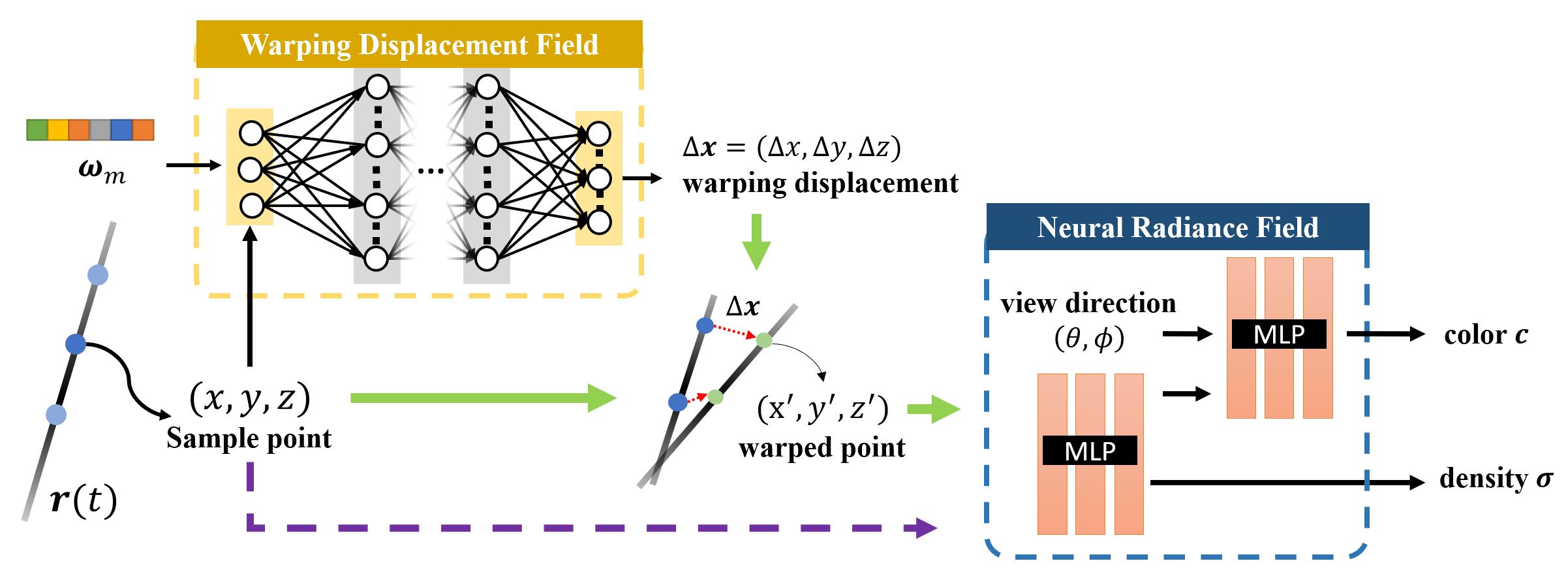}}
    		\vspace{-10pt}
	\caption{Illustration of our NeWRF rendering scheme. The latent code $\omega_m$ controls the warping field for each mirror and hence compensates calibration errors. During the novel view rendering phase, the position of the sample point is directly fed into the neural radiance field(purple dotted line).}
	\label{fig:network}
\end{figure*}

\vspace{2mm}\noindent{\bf System Calibration.} 
As illustrated in Fig.~\ref{fig:mirrot_light_path}, the rays from the target scene are captured by the camera after being reflected by the sphere mirror array. 
To enable neural scene reconstruction in a casual capture setting, we perform online calibration of our catadioptric system to restore rays in the world coordinate system from image pixels. 

To this end, we utilize the red calibration points to mark the corners of hexagons on the flat paper, which helps us to obtain the relative pose between the camera and mirror array. 
Specifically, we define this honeycomb pattern in the world coordinate system, and thus the $i$-th calibration point has a known 3D coordinate, defined as $\mathbf{X}_i = (x_i,y_i,0)^T$. 
Let $\mathbf{x}_i = (u_i,v_i)^T$ be the projection of $\mathbf{X}_i$ on the image plane.
Then, there exists a projection matrix $\mathbf{P} = \mathbf{K}[\mathbf{R}|\mathbf{t}]$ to enable the following operation:
\begin{equation}
\begin{aligned}
     \lambda \begin{bmatrix}
    \mathbf{x}_i \\ 1
    \end{bmatrix}
    =\mathbf{P}\begin{bmatrix}
    \mathbf{X}_i \\1
    \end{bmatrix}
    &= \mathbf{K}\left[\begin{array}{cccc}
    \mathbf{r}_1 , & \mathbf{r}_2 ,& \mathbf{r}_3 ,& \mathbf{t} \end{array}\right]\begin{bmatrix}
    x_i \\ y_i \\ 0 \\ 1
    \end{bmatrix}\\
    &= \mathbf{K} \left[\begin{array}{ccc}\mathbf{r}_1, & \mathbf{r}_2,   & \mathbf{t} \end{array}\right]\begin{bmatrix}
    x_i\\y_i\\1
    \end{bmatrix}\\
        &=\mathbf{H} \begin{bmatrix}
  x_i\\y_i\\1
    \end{bmatrix},
\end{aligned}
\label{eq:projection}
\end{equation}
where $\mathbf{H}$ is the homography matrix which relates the transformation between the mirror plane and the image plane; $\mathbf{K}$ is the pre-calibrated camera intrinsic matrix; $\mathbf{R}=[\mathbf{r}_1, \  \mathbf{r}_2, \ \mathbf{r}_3]$ and $\mathbf{t}$ are the world-to-camera rotation and translation, respectively. 
Note that the homography matrix $\mathbf{H}$ has $8$ degrees of freedom. 
Thus, given more than four reliable detected pairs of $\mathbf{x}_i$ and $\mathbf{X}_i$, we can estimate $\mathbf{H}$ by using the Direct Linear Transformation(DLT) algorithm. 

Furthermore, we decompose $\mathbf{H}$ into the extrinsic matrix $[\mathbf{R}|\mathbf{t}]$ using an efficient solution. 
According to the Eqn.~\ref{eq:projection}, we have the following formulation:
\begin{equation}
  \alpha \mathbf{K}^{-1}\mathbf{H} = \left[\begin{array}{ccc}\mathbf{r}_1, & \mathbf{r}_2,  & \mathbf{t} \end{array}\right].
\end{equation}
To solve the scale ambiguity, we obtain $[\mathbf{r}_1 \ \mathbf{r}_2 | \ \mathbf{t} ]$ by normalizing $\mathbf{K}^{-1}\mathbf{H}$ with the norm of its first column, so that $\|\mathbf{r}_1\|_2$ is close to one. 
Then, the initial extrinsic matrix $[\mathbf{R'}|\mathbf{t}]$ can be obtained as follows:
\begin{equation}
  [\mathbf{R'}|\mathbf{t}] = \left[\begin{array}{cccc}\mathbf{r}_1, & \mathbf{r}_2,  & \mathbf{r_1}\times \mathbf{r_2}, &  \mathbf{t} \end{array}\right].
\end{equation}
However, this solution does not ensure that $\mathbf{R'}$ will be orthogonal and have a determinant of $1$. 
Thus, we apply the orthogonal Procrustes to $\mathbf{R'} = \mathbf{U}\mathbf{\Sigma}\mathbf{V}^T$ and take $\mathbf{R} = \mathbf{U}\mathbf{V}^T$ as the final rotation matrix.


\vspace{2mm}\noindent{\bf Ray Restoration.} 
Each restored ray is a sample of the plenoptic function of the mirror array, which consists of the observation point on the mirror surface $\mathbf{o}$, the ray direction $\mathbf{d}$ and the color $\mathbf{c}$.
After the above online calibration between the mirror array and the camera in our catadioptric system, we render the geometry of the mirror array to a depth map and a normal map.
Hence we can associate the image pixels with corresponding 3D points on mirror surfaces. 
For each pixel from the mirror array, there are two rays on the light path, namely the camera ray $\mathbf{r_c}(t)$ and the ray $\mathbf{r}(t)$ we are going to restore.
$\mathbf{r_c}(t) = \mathbf{o}_c + t\mathbf{d}_c$ casts into the scene from the pixel whose image coordinate is $(u,v)$, where $\mathbf{o}_c$ is the position of the camera. 
The camera ray direction $\mathbf{d}_c$ is formulated as:
\begin{equation}
\begin{split}
\mathbf{d}_c = \frac{\mathbf{R}^T\mathbf{K}^{-1}[u \ v \ 1]^T}{\|\mathbf{K}^{-1}[u \ v \ 1]^T\|_2}.
\end{split}
\end{equation}
Then, we restore $\mathbf{r}(t) = \mathbf{o} + t\mathbf{d}$ to sample and reconstruct the captured neural scene. 
Firstly, we calculate the intersection of ray $\mathbf{r_c}(t)$ and the mirror surface, namely $\mathbf{o}$. Since we have the depth $t_d$ for the pixel, we have $\mathbf{o} = \mathbf{o}_c + t_d\mathbf{d}_c$ directly. Next, we compute the direction $\mathbf{d}$. Let $\mathbf{n}_c$ denotes the normal of $\mathbf{o}$ in camera frame. It can be obtained from the normal map. $\mathbf{n} = \mathbf{R}^T\mathbf{n}_c$ is the transformed normal in the world coordinate system. Thus $\mathbf{d}$ can be computed as in:
 
\begin{equation}
\begin{split}
\mathbf{d} = \mathbf{d}_c - 2(\mathbf{n}^T\mathbf{d}_c)\mathbf{n}.
\end{split}
\end{equation}
  
Finally, we assign the corresponding pixel color to the ray $\mathbf{r}(t)$. 
Fig.~\ref{fig:mirrot_light_path} illustrates the catadioptric light path in our system.

%
%
%

\section{Neural Portrait Rendering} \label{Sec:neural_rendering}

Here, we introduce our neural portrait rendering scheme based on the one-shot image from our flexible catadioptric system.
Specifically, a {\it Neural Warping Radiance Field} (NeWRF) is presented to restore the 3D scene and produce photo-realistic images at arbitrary viewpoints, which handles the misalignment caused by manual arrangement and coarse calibration. 
To further improve the rendering quality and training efficiency, we also propose a density regularization scheme that leverages the inherent geometry information from captured data in an effective self-supervision manner.

\subsection{Neural Warping Radiance Field}

The neural radiance field~\cite{mildenhall2020nerf} is a continuous representation for mapping each 3D point $\mathbf{x} = (x,y,z)$ and a viewing direction $\mathbf{d}=(\theta,\phi)$ to density $\sigma$ and color $\mathbf{c}=(r,g,b)$. 
Please refer to \cite{mildenhall2020nerf} for more details.
Here, we further introduce a novel warping displacement field to alleviate the influence of the misalignment. 

The warping displacement field is based on the key observation that if a mirror's position deviates from where it should be, the rays from this entire mirror are calculated wrongly and share a similar error pattern. 
Thus, our warping displacement field is conditioned on the mirror to model such error pattern implicitly, which maps each sample point from rays emitted from the same mirror to a displacement which can warp this point into a reference space. 
To this end, we assign a learnable latent code $\mathbf{\omega}_m$ that modulates the warping field to the $m$-th mirror. 
Then, our NeWRF is formulated as:
\begin{equation}
\begin{split}
F(\mathbf{x}+\Delta\mathbf{x}, \mathbf{d}) &= (\mathbf{c}, \sigma) \\
\Delta\mathbf{x} &= \Psi_{warp}(\mathbf{x}, \mathbf{\omega}_m),
\end{split}
\label{eq:NeWRF}
\end{equation}
where $\Psi_{warp}$ is the warping displacement field represented by an MLP network; $\mathbf{x}$ is a sample point from the $m$-th mirror; 
$\Delta\mathbf{x}$ is the displacement for the 3D point, while $F$ is the neural radiance field described in~\cite{mildenhall2020nerf}.  
Note that we only warp the position of the sample point, but its viewing direction remains unchanged.
As shown in Fig.~\ref{fig:network}, all points are warped into a reference space where the neural radiance field is defined, which is constrained according to the Eqn.~\ref{eq:NeWRF}. 

This may cause a global distortion between the reference space and the world coordinate system. 
Thus, we utilize the mirror in the array center as an anchor to regularize the reference space. 
Specifically, we do not apply $\Psi_{warp}$ on the sample points from the central mirror and feed them directly into $F$ instead.  
Such strategy enables that the reference space can be aligned with the world coordinate system as close as possible.

\subsection{Density Regularization}
The original NeRF~\cite{mildenhall2020nerf} takes the multilayer perceptron (MLP) training as an optimization process to regress the solution of continuous densities and colors in the space. 
However, the direction of rays in our setting is more concentrated than those multi-image ones, which means our setting lacks enough rays from the back-view or side-view to efficiently supervise the regression, especially for the density. 
It will easily cause degenerated solutions (as illustrated in Fig.~\ref{density}) and slow convergence. 
During volume rendering, sample points near the visible geometry surface should contribute most to the rendering result, and their corresponding weights are dominating.
Thus, we introduce a series of density regularization to alleviate these problems, which leverages the inherent geometry from captured data in a self-supervised manner.

\vspace{2mm}\noindent{\bf Bounding Box Sampling.} Instead of using near-far planes to sample 3D points along with rays, we pre-define a 3D bounding box in the world coordinate and only sample points inside it. 
This strategy reduces the number of points which should be mapped by our NeWRF and limits the solution space.

\vspace{2mm}\noindent{\bf Implicit Visual Hull.} We further segment the captured image into foreground and background using the Chroma Key algorithm. Sample points $\mathbb{X}_b$ located on the rays which belong to background mask pixels are in void space, and their densities should be close to zero. 
Thus, we propose a visual hull loss $\mathcal{L}_{v}$:

\begin{equation}
\begin{split}
\mathcal{L}_{v} = \frac{1}{|\mathbb{X}_b|}\sum_{\mathbf{x} \in \mathbb{X}_b}|\sigma(\mathbf{x})|^2,
\end{split}
\end{equation}
where $\sigma(\mathbf{x})$ is the density of sample point $\mathbf{x}$.

\vspace{2mm}\noindent{\bf Geometry-aware Regularization.} Unfortunately, the visual hull algorithm will produce a superfluous convex hull in front of objects in our setting. 
And our visual hull loss above is nonfunctional on this space. 
Even points inside the superfluous hull have a low-density level. It is enough to cause noise rendering results.
Thus, we introduce a self-supervision approach to regularize the density in these regions.
Given the sample points along a pixel ray $\mathbf{r}(t) = \mathbf{o} + t\mathbf{d}$, we have the parameters $t_i, i\in\{1,...,n\}$ for the $n$ sampled points. 
Then, the depth $D(r)$ is formulated as:
\begin{equation}
\begin{split}
D(\mathbf{r}) &= \sum_{i=1}^{n}W_i(1-\exp(-h(\sigma_i)\delta_i))t_i, \\
W_i &= \exp(-\sum_{j=1}^{i-1}h(\sigma_j)\delta_j),
\end{split}
\end{equation}
where $\delta_i = t_{i+1}-t_i$ is the distance between adjacent samples. 
$h(\cdot)$ is a piece-wise function as follows:
\begin{equation}
\begin{split}
h(x) = 
 \left\{
\begin{array}{ll}
      0 & x\leq \tau \\
      x & otherwise \\
\end{array} 
\right. 
\end{split}
\end{equation}
where $\tau$ is a threshold.
Note that $h(\cdot)$ filters those noises with low density in order to obtain more accurate depth values. 

During the network training, we randomly sample void point $\mathbf{x}_g$ for each ray in the certain range, where $\mathbf{x}_g = \mathbf{o} + t_g\mathbf{d}$ and $t_g \in [0, D(\mathbf{r}))$. 
Let $\mathbb{X}_g$ denotes the set of these void points of a bunch of rays. 
Then, the geometry-aware regularization loss is formulated as:
\begin{equation}
\begin{split}
\mathcal{L}_{g} = \frac{1}{|\mathbb{X}_g|}\sum_{\mathbf{x} \in \mathbb{X}_g}|\sigma(\mathbf{x})|^2.
\end{split}
\end{equation}


%





\subsection{Implementation Details}
We model both the warping displacement field $\Psi_{warp}$ and the neural radiance field $F$ as continuous functions using two separate MLPs. We train the MLP for the neural radiance field for the first three epochs individually to warm up our network design, and then two MLPs are trained together afterward.
Our network structure is illustrated in Fig.~\ref{fig:network}. 
The warping displacement network $\Psi_{warp}$ has 5 layers, 128 hidden neurons, and ReLU activation, while the activation of the last layer is removed in $\Psi_{warp}$.  
$F$ has the same structure as in the original NeRF~\cite{mildenhall2020nerf}. 

Training rays are generated from the captured image by the principle described in Sec.~\ref{Sec:capture_system}. 
We also mark the corresponding mirror index for each ray so that we can choose the corresponding warping latent code $\omega_m$ during training. 
We use 16 dimensions for the warping latent codes and optimize them through backpropagation. 
During training, we use Adam to optimize network parameters with a learning rate of $1e^{-4}$. 
The total loss $\mathcal{L}_{total}$ contains three parts: the same photometric loss $\mathcal{L}_{c}$ in the original NeRF~\cite{mildenhall2020nerf}, the visual hull loss  $\mathcal{L}_{v}$ as well as the geometry-aware regularization loss $\mathcal{L}_{g}$. 
Specifically, the total loss is formulated as:

\begin{equation}
\begin{split}
\mathcal{L}_{total} =\mathcal{L}_{c} + \lambda(\mathcal{L}_{v} + \mathcal{L}_{g}),
\end{split}
\end{equation}
where $\lambda = 10^{-2}$ in our experiments. 
Since the estimated depth $D(\mathbf{r})$ is full of noise and unreliable during the beginning of the training, 
$\tau$ is $0$ in the warm-up epochs. After that, we adjust $\tau$ progressively during training, which increases linearly from $0$ to $20$, and reaches the maximum value after five epochs. 

During inference, we remove the warping displacement field $\Psi_{warp}$ from networks and directly query radiance and densities of sample points in the reference space and adopt the volumetric integration strategy similar to \cite{mildenhall2020nerf}, so as to enable photo-realistic one-shot novel view synthesis.
%


\begin{figure*}[t]
\centering
	\centerline{\includegraphics[width=1.\textwidth]{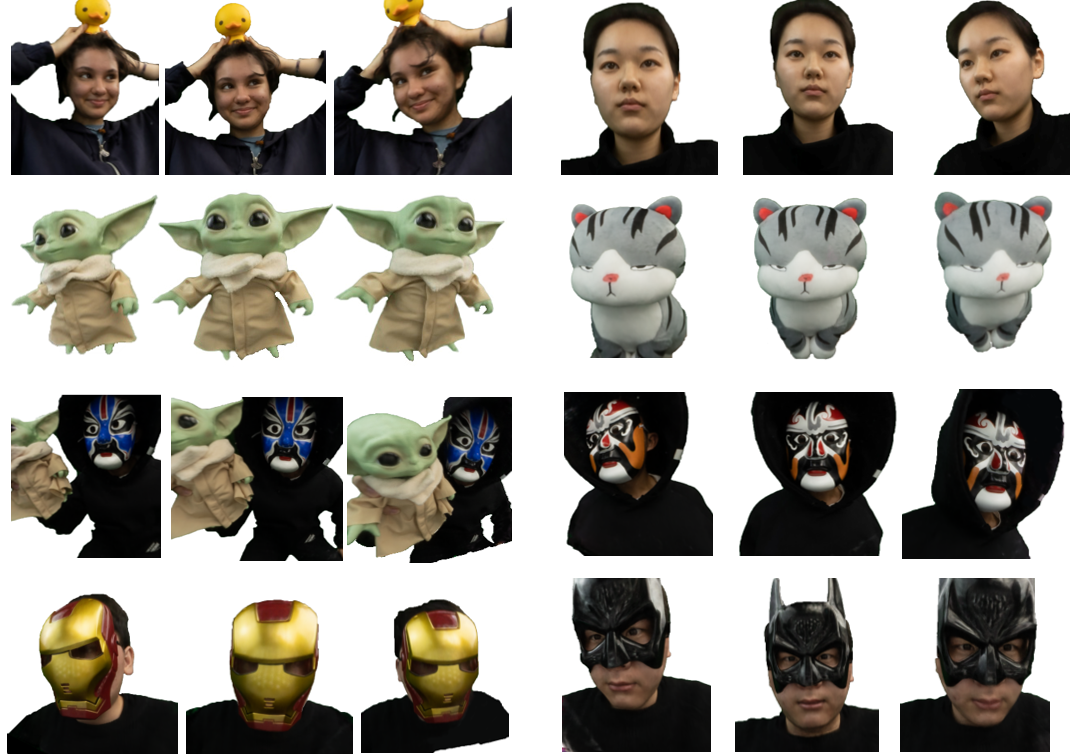}}
	\vspace{-10pt}
\caption{Several examples demonstrate our proposed system's quality and fidelity render results on the data we captured, including human portrait, objects, and human with things.}

\label{result_gallery}
\end{figure*}
	\vspace{-10pt}

\section{Experimental Results}




In this section, we evaluate the free view rendering result of our MirrorNeRF on various scenarios, followed by the comparison with other methods, both qualitatively and quantitatively. 
We provide the limitation and discussions regarding our approach in the last subsection.
Fig.~\ref{result_gallery} shows that our approaches can generate free-view high-resolution rendering with high-quality details. 
We train our model on Nvidia GeForce RTX3090 GPU for 10 hours with 4000 rays per batch.
Each ray uses 96 samples in coarse volume and 32 additional adaptive samples in fine volume for volumetric integration. 
We render images with the output resolution of $1200\times 900$ in our experiments. Our approach takes about 14 seconds to render each image.

\subsection{Evaluate}

\begin{figure}[t]
\centering
	\centerline{\includegraphics[width=.5\textwidth]{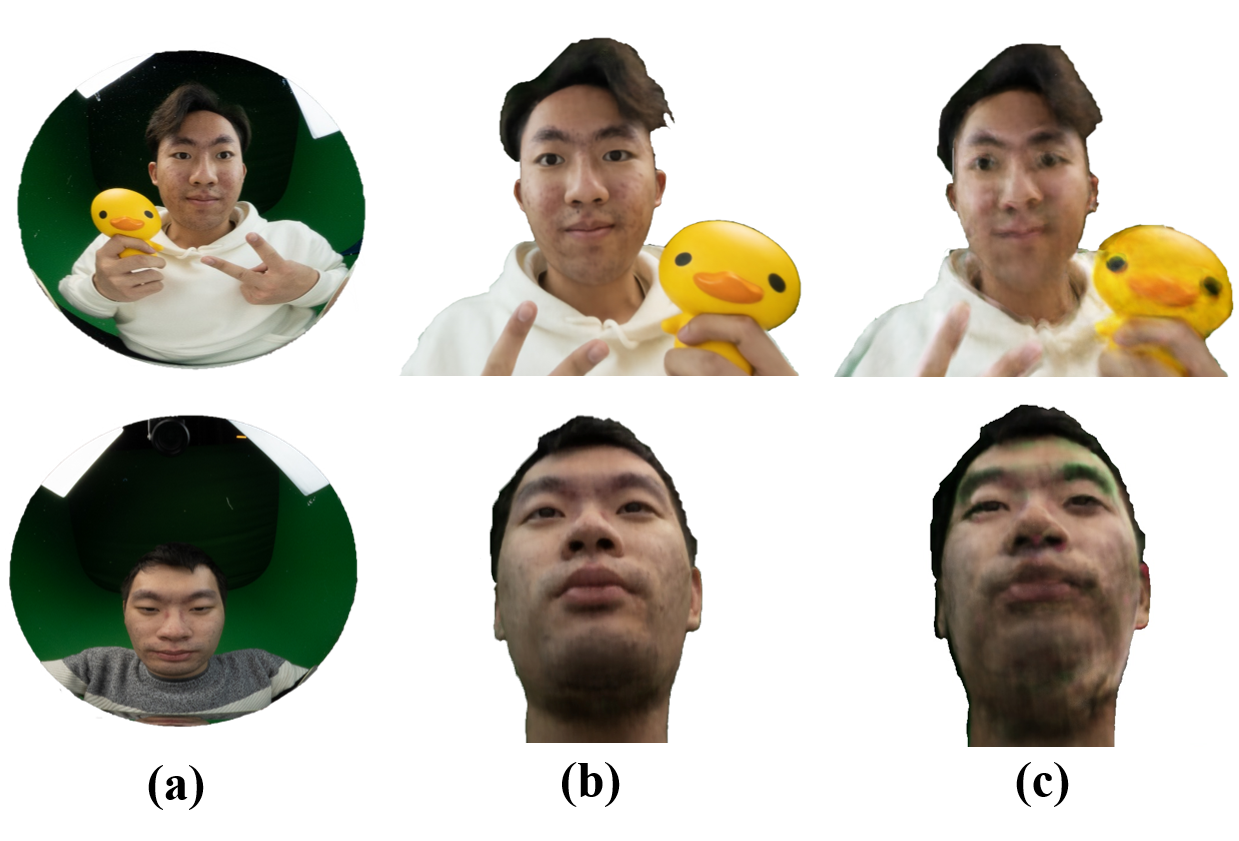}}
		\vspace{-10pt}
\caption{Demonstrations of neural warping field evaluation. (a) captured reference image from one mirror in the array; (b) results with warping module; (c) results without warping module.}
\label{deform}
\end{figure}
\vspace{1mm}\noindent{\bf Neural Warping Field Evaluation.}

\begin{figure}[t]
\centering
\centerline{\includegraphics[width=.5\textwidth]{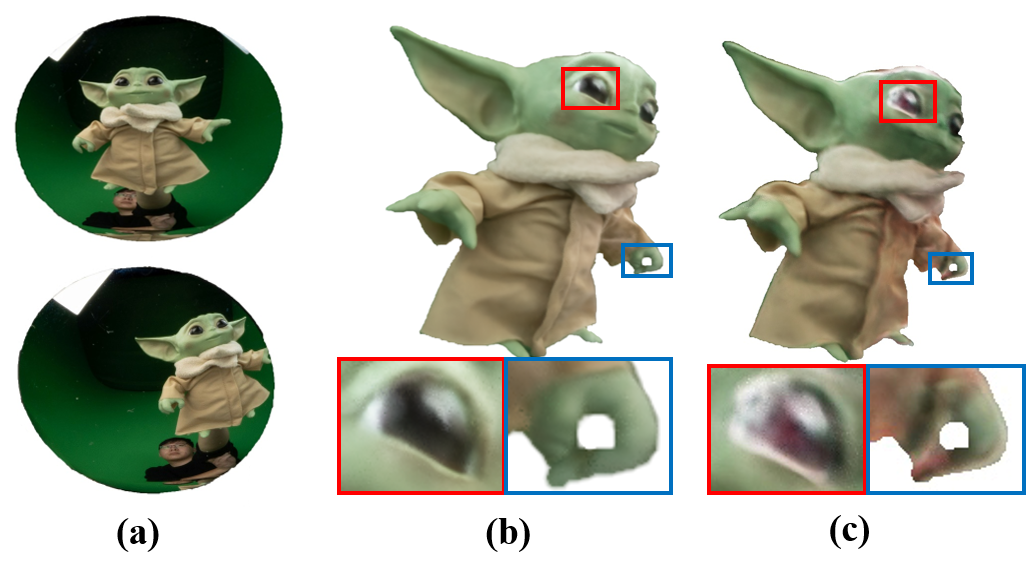}}
	\vspace{-10pt}
\caption{Demonstration of density regularization evaluation. (a) captured reference image from 2 mirrors in the array; (b) results with density regularization; (c) results without density regularization.}
\label{density}
\end{figure}

\begin{figure*}[t]
\centering
\centerline{\includegraphics[width=\textwidth]{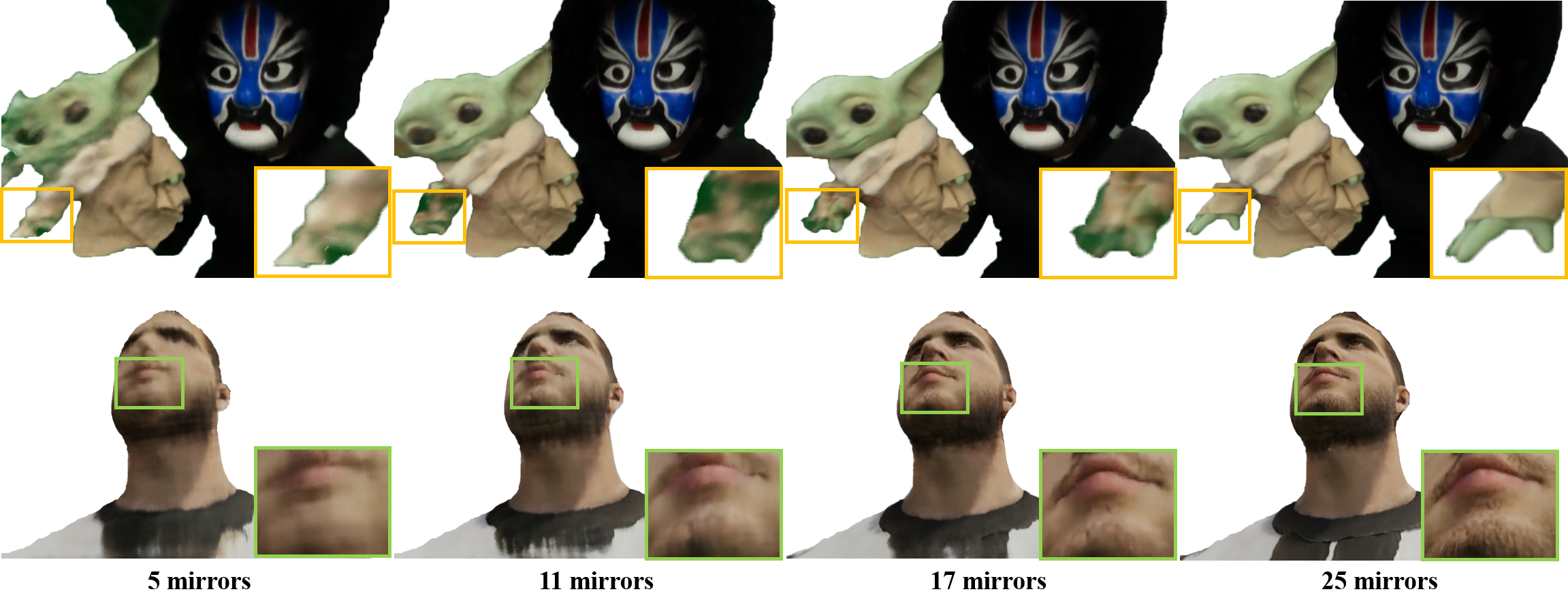}}
	\vspace{-10pt}
\caption{Qualitative evaluation of the number of mirrors on both real data and synthetic data. Geometry becomes incomplete, and details are lost when decreasing the number of mirrors.}
\label{Number_Mirror}
\end{figure*}
To evaluate our neural warping radiance field, we train our network with/without warping displacement field module $\Psi_{warp}$. 
As shown in Fig.~\ref{deform}, results without warping field exhibit strong blurring, while others with warping field alleviate this phenomenon and produce photo-realistic images.  
%
%
This evaluation shows that calibration errors caused by manual placement have greatly influenced the neural radiance field regression and lead to poor results. Original NeRF does not have the ability to correct these errors. Our NeWRF with warping displacement field module successfully tackles this problem.     

%
%
%

\vspace{1mm}\noindent{\bf Density Regularization Evaluation.}
We further evaluate our density regularization. 
As shown in Fig.~\ref{density}, the rendered image appears fine details with density regularization. 
We can observe some noises in the result images without density regularization, especially in some areas such as hands and eyes. The regression of neural radiance field in these areas will easily be trapped in a local minimum and causes noises without density regularization. On the contrary, our density regularization scheme uses the geometric knowledge priors learned from training to optimize the result.

%
%

\begin{figure}[t]
\centering
\centerline{\includegraphics[width=.5\textwidth]{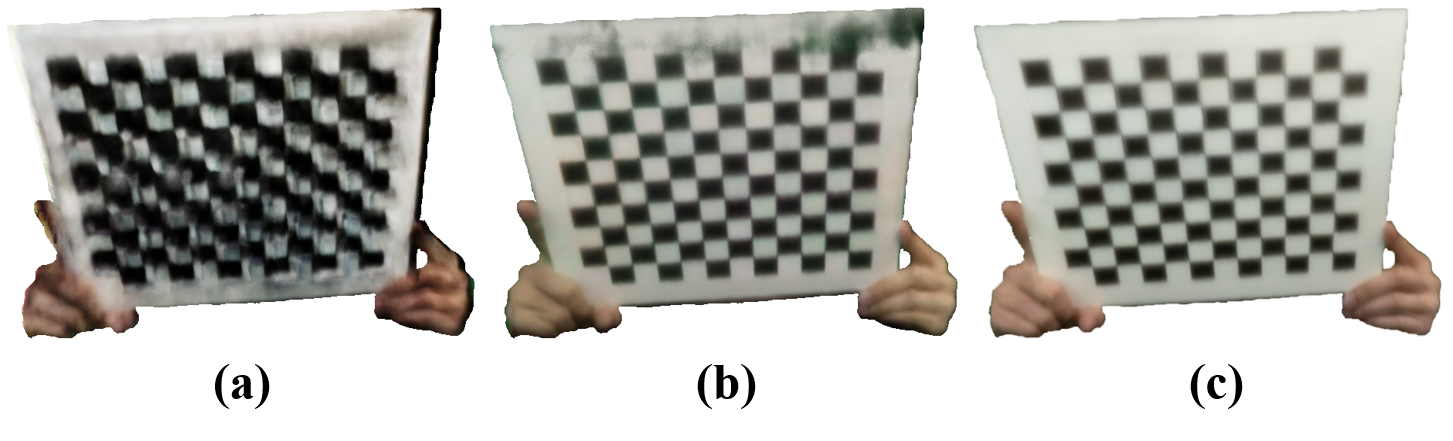}}
	\vspace{-10pt}
\caption{Qualitative evaluation of our method to reconstruct the same checkerboard. (a) w/o neural warping. (b) w/o density regularization. (c) ours.}
\label{checkboard}
\end{figure}
\begin{table}[t]
\centering
\caption{Quantitative evaluation on reconstructing the same checkerboard. 
}
\vspace{-10pt}
\begin{tabular}{c|cc}
\hline
  Methods           & \begin{tabular}[c]{@{}c@{}}Detection\\ Success Rate\end{tabular} & \begin{tabular}[c]{@{}c@{}}Mean\\ Reprojection Error\end{tabular} \\ \hline
w/o Neural warping        & 0.7595                                                           & 0.1916                                                            \\
w/o Density Reg.          & 0.8931                                                           & 0.1249                                                            \\
Ours                     & \textbf{0.9656}                                                  & \textbf{0.1197}                                                   \\
 \hline
\end{tabular}

\label{table:checkboard}
\end{table}

\begin{table}[t]
\centering
\caption{Quantitative comparison on synthetic data.}
\vspace{-10pt}
\begin{tabular}{@{}c|ccc@{}}
\toprule
                   & PSNR    $\uparrow$        & SSIM    $\uparrow$        & LPIPS   $\downarrow$       \\ \midrule

w/o Neural warping & 22.64          & .8838          & .1547          \\
w/o Density Reg.    & 24.29          & .9044          & .0861          \\
Ours               & \textbf{24.82} & \textbf{.9070} & \textbf{.0854} \\ \bottomrule
\end{tabular}

\label{Eval_Synthetic}
\end{table}

We also quantitatively analyze the rendering quality on both real and synthetic data. 
As for real data, we train the same checkerboard data in three ways and render each of them in 262 views. Fig.~\ref{checkboard} illustrates one image of them. Both the model in Fig.~\ref{checkboard}(a) and Fig.~\ref{checkboard}(a) appear unpleasant artifacts. In contrast, our complete model reconstructs the checkerboard in detail and with less noise.
To quantify the rendering quality, We define two metrics. 
The average Detection Success Rate measures the rendering quality of the reconstructed corner.
For each view, only when all the corners of the checkerboard are detected will it be considered successful. The Mean Reprojection Error measures the geometric correctness of reconstruction. We first obtain the checkerboard pose using all checkerboard corners. Then, we project the real checkerboard corners to the corresponding view to evaluate the reprojection error.
As shown in Tab.~\ref{table:checkboard}, our complete model achieves the best result in both detecting corners and reprojection. 

As for synthetic data, we add Gaussian noise on the mirror positions to simulate the actual situation. And we all render the same 97 novel views for each method and compare them to the ground truth. 
Tab.~\ref{Eval_Synthetic} summarizes the effects of our model components.
Our full model performs the best in all metrics for evaluation: Peak Signal-to-Noise Ratio (PSNR), Structural Similarity (SSIM) \cite{SSIM} and
Learned Perceptual Image Patch Similarity (LPIPS) \cite{LPIPS}.


\begin{figure}[t]
\centering
\centerline{\includegraphics[width=.48\textwidth]{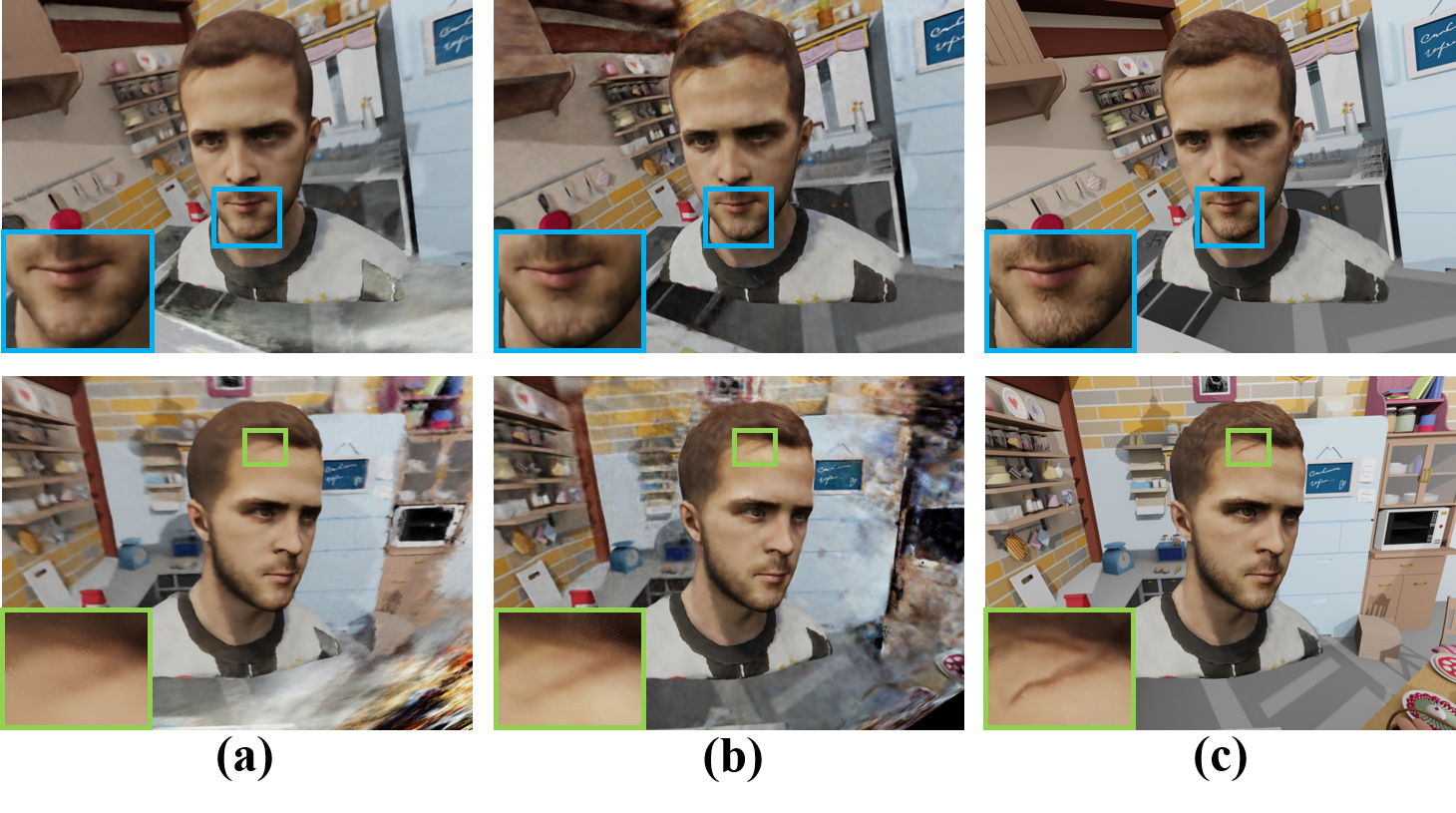}}
	\vspace{-10pt}
\caption{Qualitative evaluation of the scene with textured background and parallax. (a) NDC sampling  w/o density regularization. (b) Ours w/o density regularization. (c) Ours.}
\label{fig:Eval_Synthetic_BG}   
\end{figure}



\begin{table}[t]
\centering
\caption{Quantitative evaluation of the scene with textured background and parallax.}
\vspace{-10pt}
\begin{tabular}{@{}c|ccc@{}}
\toprule
                   & PSNR    $\uparrow$        & SSIM    $\uparrow$        & LPIPS   $\downarrow$       \\ \midrule

NDC w/o Reg. & 23.19     & .8979    & .1055   \\
Ours w/o Reg.    & 24.16          & .9019          & .0956          \\
Ours               & \textbf{24.41} & \textbf{.9049} & \textbf{.0896} \\ \bottomrule
\end{tabular}
\label{table:Eval_Synthetic_BG}
\end{table}

We further evaluate the influence when the background is not a green screen. 
%
%
In this scenario, We first evaluate our method using the NDC sampling strategy, which is proposed in ~\cite{mildenhall2020nerf} to handle large depth.  As shown in Fig.~\ref{fig:Eval_Synthetic_BG}(a), NDC helps reconstruct the scene with background, but it decreases the rendering quality of the portrait.
Our approach only without the regularization in Fig.~\ref{fig:Eval_Synthetic_BG}(b) also reduces the foreground's quality while learning the background and the foreground at the same time.
In contrast, our entire pipeline in Fig.~\ref{fig:Eval_Synthetic_BG}(c) enables more photo-realistic rendering of the foreground
and further compositing purposes. 
We provide the corresponding quantitative analysis in Table.~\ref{table:Eval_Synthetic_BG}. 
It shows the effectiveness of our density regularization for reconstructing the foreground, especially for handling our unique catadioptric setting.
Note that in Figure~\ref{fig:Eval_Synthetic_BG}(b)(c), we expand the bounding box during training and rendering to cover the background.
While in our current regularization design with a green screen, we can only sample rays from the foreground bounding box to avoid unnecessary sampling and achieve better results.

\begin{table}[t]
\centering
\caption{Quantitative evaluation on different number of mirrors. 
}
\vspace{-10pt}
\begin{tabular}{@{}c|ccc@{}}
\toprule
\# Mirrors & PSNR   $\uparrow$        & SSIM    $\uparrow$        & LPIPS    $\downarrow$       \\ \midrule
5          & 21.72          & .9009          & .1356          \\
11         & 22.38          & .9001          & .1159          \\
17         & 23.15          & .9030          & .0948          \\
25         & \textbf{24.82} & \textbf{.9070} & \textbf{.0854} \\ \bottomrule
\end{tabular}

\label{table:num_mirror}
\end{table}

\begin{figure*}[t]
\centering
\centerline{\includegraphics[width=1\textwidth]{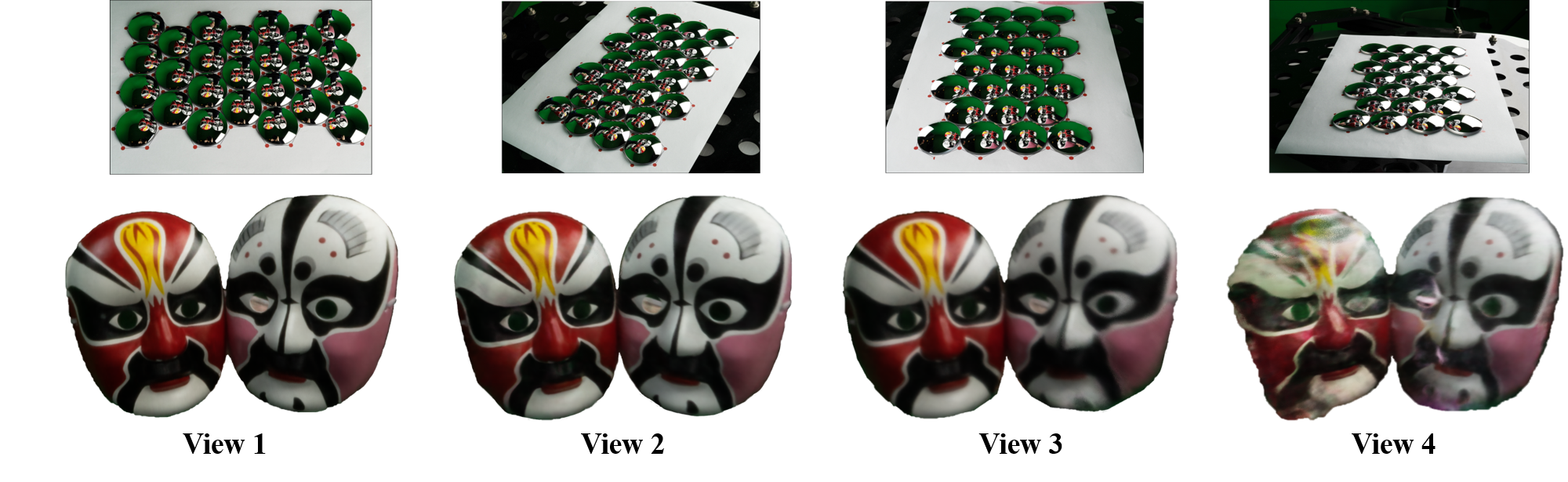}}
	\vspace{-10pt}
\caption{Evaluation of different camera positions. Our catadioptric system is unstructured, which can capture the image of the mirror array from various view angles of a handheld camera.}
\label{Different_view}
\end{figure*}

\vspace{1mm}\noindent{\bf Mirror Number Evaluation.}
To evaluate the influence of the number of mirrors,  we only use the reflected rays of selected 5, 11, 17, and 25 mirrors, respectively. 
As Fig.~\ref{Number_Mirror} shows, the reconstruction result suffers from geometry and rendering artifacts at the non-central part of the scene when decreasing the mirror numbers.
Quantitatively, Tab.~\ref{table:num_mirror} shows that using 25 mirrors outperforms the best in all scenarios.
%

\vspace{1mm}\noindent{\bf Camera Position Evaluation.}
To demonstrate that our system is unstructured, we reconstruct the same facial mask by capturing the data in 4 different views using a handheld camera. 
The first row in Fig.~\ref{Different_view} is the raw captured data. 
View 1 is the viewpoint we use most in our system, where most of the pixels are the reflected rays of mirrors.
View 2 and View 3 are at a certain angle to our catadioptric system. View 4 is from a really oblique perspective.
From the second row in Fig.~\ref{Different_view}, our system can achieve similar results in normal viewing angles, for example, View 1, View 2, and View3. 
In the skewed viewpoint like View 4, the performance drops and some artifacts appears.

\begin{figure}[t]
\centering
\centerline{\includegraphics[width=.5\textwidth]{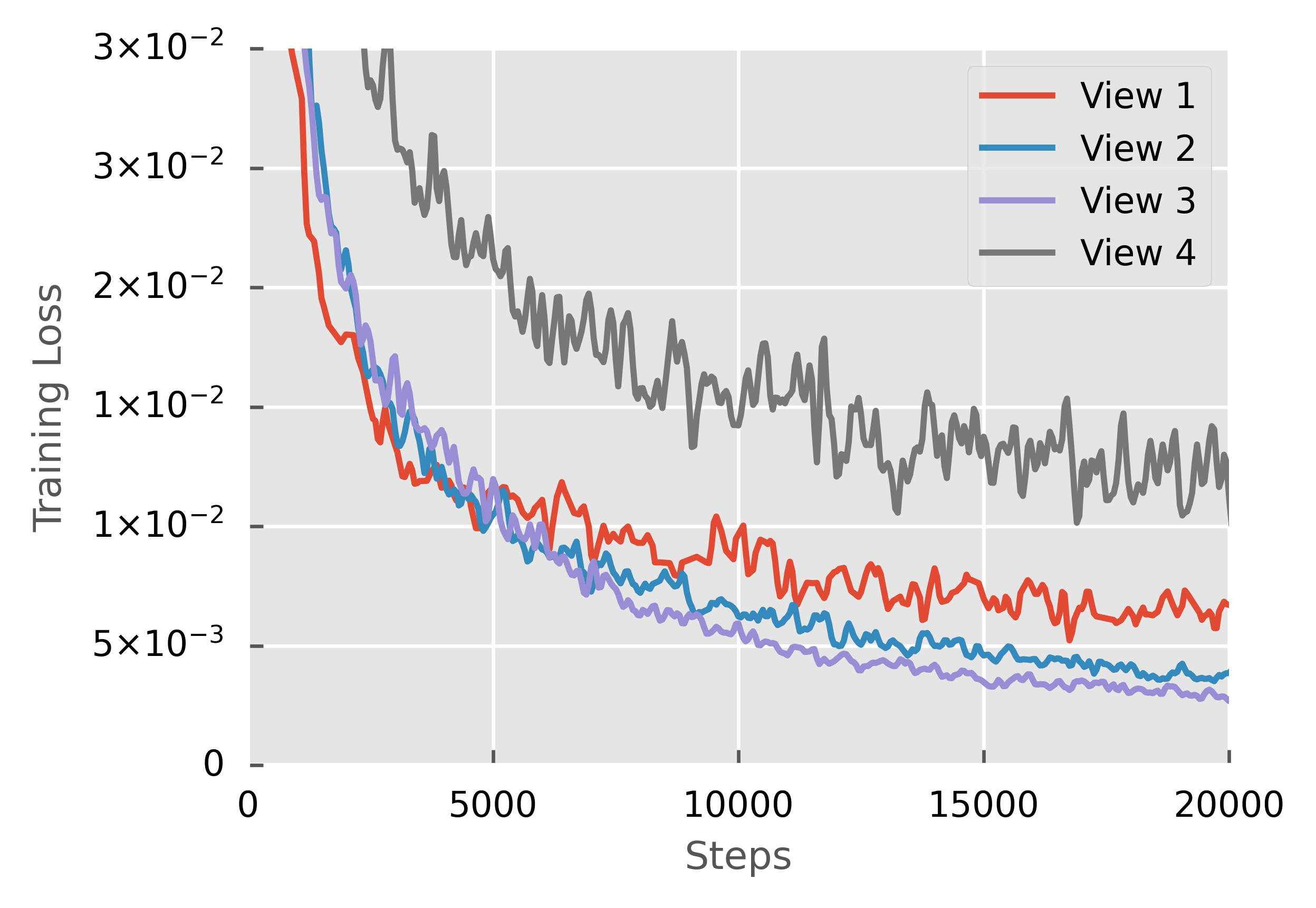}}
	\vspace{-10pt}
\caption{Training loss for different camera positions.}
\label{Different_view_curve}
\end{figure}

Fig.~\ref{Different_view_curve} are curves of the training loss for four different views.
View 1, View 2, and View 3 coverage similarly, while the fourth one is slower and has a higher loss. 
It quantitatively reveals that our system can achieve similar performance in most capturing angles and has some limitations in some skewed angles.


\begin{figure*}[t]
\centering
\centerline{\includegraphics[width=\textwidth]{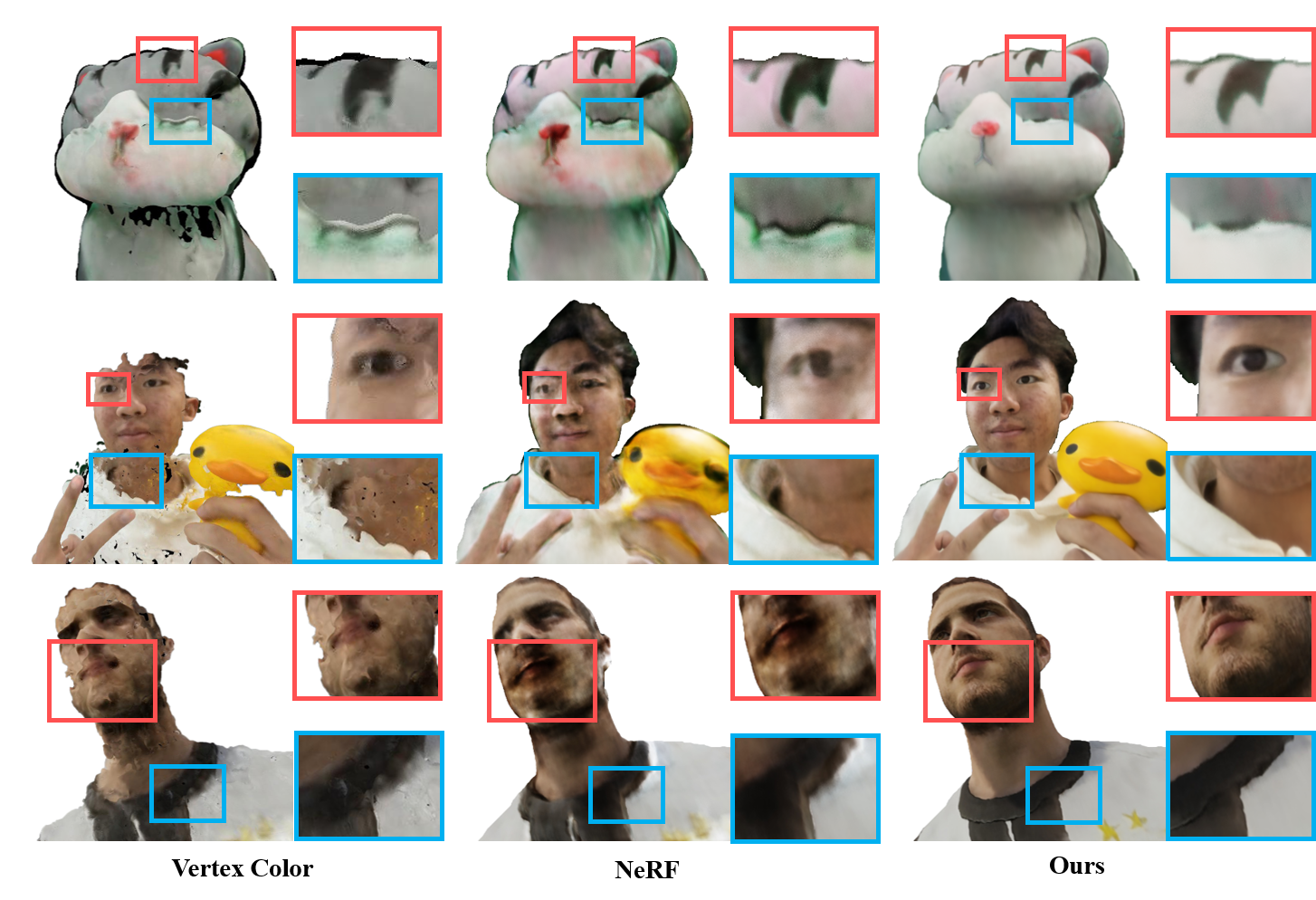}}
	\vspace{-10pt}
\caption{Qualitative comparison on both the real (the first two rows) and synthetic data (the third row). From left to right: the results of the per-vertex color scheme via ray tracing, the original NeRF and ours, respectively.}
\label{comp}
\end{figure*}
\begin{figure}[t]
\centering
\centerline{\includegraphics[width=.5\textwidth]{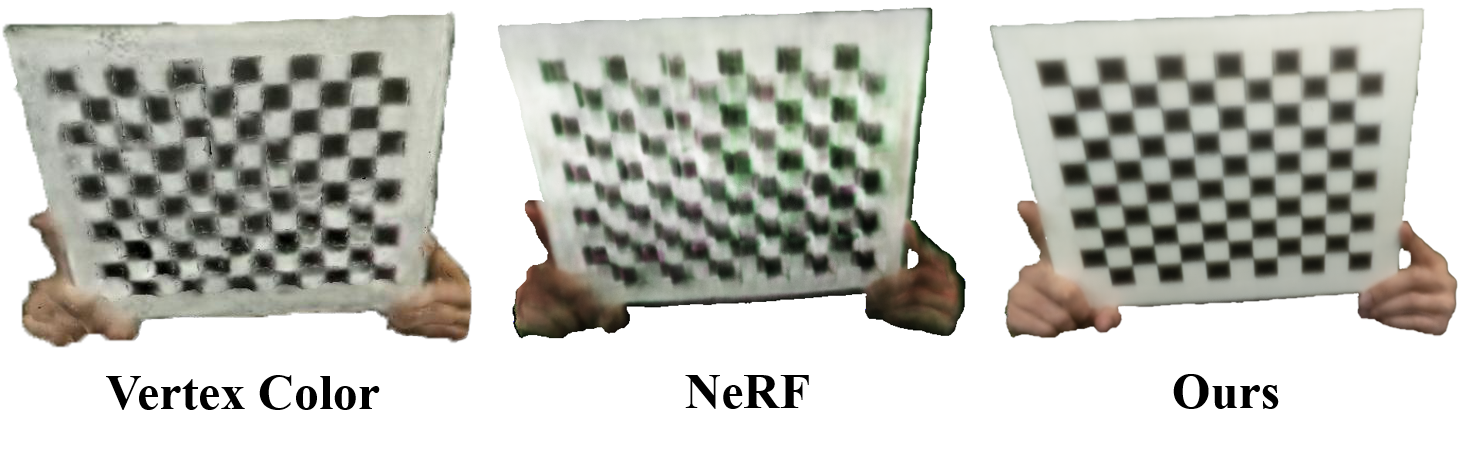}}
	\vspace{-10pt}
\caption{Qualitative evaluation of our method to reconstruct the same checkerboard. From left to right: the results of the per-vertex color scheme via ray tracing, the original NeRF and ours, respectively.}
\label{checkboard2}
\end{figure}

\subsection{Comparison}

We compare our complete model with per-vertex color and origin NeRF. 
To get the per-vertex color, we use ray tracing to query the corresponding pixel color.
Its geometry proxy is extracted from the original NeRF using marching cube with a threshold to filter the geometry noise.

As illustrated in Fig.~\ref{comp}, our method achieves significantly more photo-realistic novel view rendering results than other methods. 

To quantitatively compare our method to others, we similarly reconstruct the same checkerboard in Fig.~\ref{checkboard2}, calculate the Average Detection Success Rate  and  Mean  Re-projection  Error in Tab.~\ref{table:checkboard_comp} using three methods. Also, Tab.~\ref{Comp_Synthetic} shows the performance of the three on synthetic data with noise on positions of mirrors by comparing PSNR, SSIM, and LPIPS. Our method surpasses others in all indicators with distinct differences. Vertex color greatly relies on the geometry proxy. Without an accurate geometry and enough vertices of the 3D model, it cannot provide a high-quality rendering. Also, it cannot support view-dependent effects. On the other hand, origin NeRF suffers from noises in the density field if the view directions are not diverse enough. It is also unable to handle the misalignment due to the inaccurate position of mirrors. In contrast, our method is capable of reconstructing the scene with high quality in our catadioptric setting.
 

\begin{table}[t]
\centering
\caption{Quantitative comparison on reconstructing the same checkerboard. 
}
\vspace{-10pt}
\begin{tabular}{c|cc}
\hline
  Methods           & \begin{tabular}[c]{@{}c@{}}Detection\\ Success Rate\end{tabular} & \begin{tabular}[c]{@{}c@{}}Mean\\ Reprojection Error\end{tabular} \\ \hline
Vertex Color           &  0.1870             &  0.6116         \\
NeRF                   & 0.7061        & 0.2104 \\
Ours                     & \textbf{0.9656}                                                  & \textbf{0.1197}                                                   \\
 \hline
\end{tabular}

\label{table:checkboard_comp}
\end{table}

\begin{table}[t]
\centering
\caption{Quantitative comparison on synthetic data.}
\vspace{-10pt}
\begin{tabular}{@{}c|ccc@{}}
\toprule
                  & PSNR    $\uparrow$        & SSIM    $\uparrow$        & LPIPS   $\downarrow$       \\ \midrule
Vertex Color       & 18.64          & .8456          & .1842          \\
NeRF               & 21.79          & .8833          & .1368          \\ 
Ours               & \textbf{24.82} & \textbf{.9070} & \textbf{.0854} \\ \bottomrule
\end{tabular}

\label{Comp_Synthetic}
\end{table}

\subsection{Limitation and Discussion}
As the first trial to explore the problem of combining neural radiance field with catadioptric imaging to reconstruct and render a scene in a convenient capture setting, the proposed MirrorNeRF still owns limitations as follows. 

First, our system requires that all the mirrors are almost on a plane. 
Thus, our method can only reconstruct the front view of the human or object. 
One of our further steps is to make our mirror arrangement unstructured. 
If the mirrors are placing around the human portrait or object, for example, in a sphere structure, the full body of the human or object can be reconstructed.  
Our method is also restricted to the number of distinct reflected light rays from the scene to be reconstructed. 
The high-resolution camera, number of mirrors, and not skewed viewing angle are used to provide enough light rays such that the rendering result has high fidelity even in details. 
It is interesting that the requirements of our system can be reduced to portable. We expect that our system will enable everyone to reconstruct anything they want anytime and anywhere by his or her cellphone and just several mirrors in the pocket.
Besides, it still needs hours to train for a static scene. 
Although our system is suitable for video-rate image capturing,
it is hard to reconstruct a dynamic scene in a short time.
It’s a promising direction to use some neural prior on human portrait or the kind of objects we will reconstruct or some other techniques to speed up training without decreasing the performance. Another direction is to enable our neural warping field to be deform-able in time series, so the dynamic scene is recorded in our radiance field.

\section{Conclusion}
We have presented the first approach to combine neural radiance field with catadioptric imaging, which enables one-shot, photo-realistic and free-viewpoint human portrait reconstruction and rendering.
Our light-weight catadioptric system design enables diverse ray sampling in the 3D space, which is low-cost, easily deployed and supports casual capture.
Our novel neural warping radiance field enables implicit compensation of the misalignment due to our flexible system
setting via a continuous displacement field learning.
Our self-supervised density regularization scheme further improves the training efficiency and provides more effective density supervision for higher rendering quality.
Our experimental results demonstrate the effectiveness of MirrorNeRF for compelling one-shot neural portrait rendering in various challenging scenarios, which compares favorably to the state-of-the-arts.
Given the aforementioned distinctiveness, we believe that our approach is a critical step to enable conveniently and  
photo-realistic human portrait modeling, with many potential applications in VR/AR like gaming, entertainment and immersive telepresence.

\ifpeerreview \else
\section*{Acknowledgments}
This work was supported by NSFC (grant nos. 61976138 and 61977047), and STCSM (2015F0203-000-06).
\fi

\bibliographystyle{IEEEtran}
\bibliography{iccp21_mirrorNerf}

\end{document}